%% file: main.tex
\documentclass[sigconf,authorversion,screen]{acmart} 

\copyrightyear{2022}
\acmYear{2022}
\setcopyright{rightsretained}
\acmConference[KDD '22] {Proceedings of the 28th ACM SIGKDD Conference on Knowledge Discovery and Data Mining}{August 14--18, 2022}{Washington, DC, USA.}
\acmBooktitle{Proceedings of the 28th ACM SIGKDD Conference on Knowledge Discovery and Data Mining (KDD '22), August 14--18, 2022, Washington, DC, USA}
\acmPrice{}
\acmISBN{978-1-4503-9385-0/22/08} 
\acmDOI{10.1145/3534678.3539466}

\input{commands} 
\input{notation} 

\newcites{appendix}{Appendix references}

\begin{document}

\title{On Missing Labels, Long-tails and Propensities in Extreme Multi-label Classification}


\author{Erik Schultheis}
\affiliation{%
  \institution{Aalto University}
  \city{Helsinki}
  \country{Finland}
}
\email{erik.schultheis@aalto.fi}

\author{Marek Wydmuch}
\affiliation{%
  \institution{Poznan University of Technology}
  \city{Poznan}
  \country{Poland}
}
\email{mwydmuch@cs.put.poznan.pl}

\author{Rohit Babbar}
\affiliation{%
  \institution{Aalto University}
  \city{Helsinki}
  \country{Finland}
}
\email{rohit.babbar@aalto.fi}

\author{Krzysztof Dembczyński}
\affiliation{%
  \institution{Yahoo! Research}
  \city{New York}
  \country{USA}
}
\additionalaffiliation{
  \institution{Poznan University of Technology}
  \city{Poznan}
  \country{Poland}
}
\email{kdembczynski@cs.put.poznan.pl}

\renewcommand{\shortauthors}{Erik Schultheis et al.}



\begin{abstract}
The propensity model introduced by~\citet{Jain_et_al_2016} has become a standard approach for dealing with missing and long-tail labels in extreme multi-label classification (XMLC). In this paper, we critically revise this approach showing that despite its theoretical soundness, its application in contemporary XMLC works is debatable. We exhaustively discuss the flaws of the propensity-based approach, and present several recipes, some of them related to solutions used in search engines and recommender systems, that we believe constitute promising alternatives to be followed in XMLC.
\end{abstract}

\maketitle

\section{Introduction}
\input{1-intro}

\section{Problem statement}
\label{sec:problem_statement}
We first define the problem of XMLC, 
then the problem of missing labels, 
and finally the problem of long-tail labels. 

\subsection{Extreme multi-label classification}
\label{subsec:xmlc}
\input{2-xmlc}

\section{Critical view on the current approach to sparse labels in XMLC}
\label{sec:critical-view-on-propensities}

In this section we present an overview on the current state of addressing the
long-tail and missing-labels problems in XMLC. This is in large parts based on
the work of \citet{Jain_et_al_2016}, so we start by a recap of their findings.

\subsection{Current approach to missing labels and long tails}
\input{3-critical-view}

\section{Recipes to follow}
\label{sec:recipes-to-follow}

\input{4-recipes}

\section{Conclusions}

Despite our critical comments regarding~\citep{Jain_et_al_2016}, 
we still appreciate this contribution.
It was the first paper that brought direct attention to the problem of missing and long-tail labels in XMLC.
The original theoretical results concerning the propensity model have motivated a lot of research in this direction.
Nevertheless, we believe that missing labels and long-tail labels are rather orthogonal problems 
that should be solved with different tools.
Obviously, labels gone missing may cause labels to be sparse, 
but it does not mean that a blind propensity model may solve any of these problems.

\begin{acks}
Academy of Finland grant: Decision No. 347707. Computational experiments have been performed in Poznan Supercomputing and Networking Center.
\end{acks}

\bibliographystyle{ACM-Reference-Format}
\balance
\bibliography{references_short}

\appendix
\input{5-appendix}

\bibliographystyleappendix{ACM-Reference-Format}
\balance
\bibliographyappendix{references_short}

\end{document}

%% file: commands.tex
\usepackage[utf8]{inputenc} 
\usepackage[T1]{fontenc}    
\usepackage[english]{babel}

\usepackage{microtype}      
\usepackage{siunitx}

\usepackage{hyperref}       
\usepackage{url}            
\usepackage{booktabs}       

\usepackage{amssymb}

\usepackage{subcaption}

\usepackage{amsthm}
\usepackage{amssymb}
\usepackage{amsmath}
\usepackage{dsfont}
\usepackage{natbib}

\usepackage{algorithm}
\usepackage[noend]{algpseudocode}

\usepackage{mathtools} 
\usepackage{stmaryrd}
\usepackage{multirow}
\usepackage{makecell}

\usepackage{thmtools} 
\usepackage{thm-restate}
\usepackage{balance}
\usepackage{booktabs}
\usepackage{rotating}
\usepackage{multibib}

\usepackage{tikz}
\usepackage{pgfplots}
\usepackage{pgfplotstable}
\usetikzlibrary{positioning}

\usepackage{comment}

\usepackage{adjustbox}

\makeatletter

\newcommand*{\eifstartswith}{\@expandtwoargs\ifstartswith}
\newcommand*{\ifstartswith}[2]{%
  \if\@car#1.\@nil\@car#2.\@nil
    \expandafter\@firstoftwo
  \else
    \expandafter\@secondoftwo
  \fi}

\theoremstyle{remark}

\theoremstyle{definition}

\renewcommand{\vec}[1]{\boldsymbol{#1}}

\newcommand{\bx}{\vec{x}}

\newcommand{\calY}{\mathcal{Y}}

\algnewcommand{\IIf}[1]{\State\algorithmicif\ #1\ \algorithmicthen}
\algnewcommand{\IElse}[1]{\State\algorithmicelse\ #1\ }
\algnewcommand{\IElseIf}[1]{\State\algorithmicelse \algorithmicif\ #1\  \algorithmicthen}
\algnewcommand{\EndIIf}{\unskip\ \algorithmicend\ \algorithmicif}
\algnewcommand{\IfThen}[2]{\State\algorithmicif\ #1\ \algorithmicthen\ #2\ }
\algnewcommand{\ForDo}[2]{\State\algorithmicfor\ #1\ \algorithmicdo\ #2\ }

\newcommand{\given}{\, | \,}

\DeclareMathOperator*{\argmin}{\arg \min}



%% file: notation.tex
\DeclareMathOperator{\expectsymb}{\mathds{E}}

\DeclareMathOperator{\probsymb}{\mathds{P}}
\DeclareMathOperator{\risksymb}{Risk}
\NewDocumentCommand{\indsymb}{}{\mathds{1}}
\NewDocumentCommand{\reals}{}{\mathds{R}}

\newcommand{\nnreals}{\reals_{\geq0}}

\newcommand{\numlabels}{m}
\newcommand{\numinstances}{n}
\newcommand{\propensity}{p}
\newcommand{\trainprop}{p^{\text{tr}}}
\newcommand{\estimtrainprop}{\hat{p}^{\text{tr}}}
\newcommand{\estimprop}{\hat{p}}
\newcommand{\ctprop}{p^{\text{c}}}

\newcommand{\instancespace}{\mathcal{X}}
\newcommand{\rinstance}{X}
\newcommand{\sinstance}{x}

\newcommand{\labelspace}{\mathcal{Y}}

\newcommand{\rlabels}{\vec{Y}}
\newcommand{\slabels}{\vec{y}}
\newcommand{\rblabel}{Y}
\newcommand{\sblabel}{y}
\newcommand{\rolabels}{\tilde{\vec{Y}}}
\newcommand{\solabels}{\tilde{\vec{y}}}
\newcommand{\rolabel}{\tilde{Y}}
\newcommand{\solabel}{\tilde{y}}

\newcommand{\masks}{\vec{M}}
\newcommand{\mask}{M}

\newcommand{\ocount}{\tilde{n}}

\newcommand{\rpreds}{\hat{\vec{Y}}}
\newcommand{\spreds}{\hat{\vec{y}}}

\newcommand{\spred}{\hat{y}}

\newcommand{\hypothesis}{h}
\newcommand{\bayescls}{h^*}
\newcommand{\lossfn}{\ell}
\newcommand{\ubloss}{\tilde{\ell}}
\newcommand{\taskloss}{\ell_{\text{task}}}

\newcommand{\labeldist}{\eta}
\newcommand{\olabeldist}{\tilde{\eta}}

\newcommand{\trainlabelprior}{\hat{\pi}^{\text{tr}}}
\newcommand{\vallabelprior}{\hat{\pi}^{\text{val}}}
\newcommand{\olabelprior}{\tilde \pi}
\newcommand{\hnoisylabelprior}{\hat{\olabelprior}}

\newcommand{\srelevantlabels}{\mathcal{L}}

\newcommand{\xmlcpropensitymodel}{\text{JPV}}

\NewDocumentCommand{\relevantlabels}{m}{\srelevantlabels(#1)}
\NewDocumentCommand{\set}{m}{\{#1\}}

\NewDocumentCommand{\wrapbrackets}{m}{\left[ #1 \right]}

\NewDocumentCommand{\expectation}{e{_}o}{
    \def\decorated{\expectsymb\IfNoValueTF{#1}{}{_{#1}}}
    \IfNoValueTF{#2}{\decorated}{\decorated\!\wrapbrackets{#2}}
}

\NewDocumentCommand{\probability}{o}{
    \IfNoValueTF{#1}{\probsymb}{\probsymb\!\wrapbrackets{#1}}
}

\NewDocumentCommand{\indicator}{o}{
    \IfNoValueTF{#1}{\indsymb}{\indsymb\!\wrapbrackets{#1}}
}

\NewDocumentCommand{\risk}{o}{
    \IfNoValueTF{#1}{\risksymb}{\indsymb\!\wrapbrackets{#1}}
}

\NewDocumentCommand{\defmap}{mmm}{\ensuremath{#1: #2 \longrightarrow #3}}

\NewDocumentCommand{\precision}{o}{
    \IfNoValueTF{#1}{\operatorname{P}}{\operatorname{P@}\!#1}
}
\NewDocumentCommand{\recall}{o}{
    \IfNoValueTF{#1}{\operatorname{R}}{\operatorname{R@}\!#1}
}
\NewDocumentCommand{\ndcg}{o}{
    \IfNoValueTF{#1}{\operatorname{nDCG}}{\operatorname{nDCG@}\!#1}
}
\NewDocumentCommand{\psprecision}{o}{
    \IfNoValueTF{#1}{\operatorname{PSP}}{\operatorname{PSP@}\!#1}
}
\NewDocumentCommand{\psrecall}{o}{
    \IfNoValueTF{#1}{\operatorname{PSR}}{\operatorname{PSR@}\!#1}
}
\NewDocumentCommand{\psndcg}{o}{
    \IfNoValueTF{#1}{\operatorname{PSnDCG}}{\operatorname{PSnDCG@}\!#1}
}
\NewDocumentCommand{\abandon}{o}{
    \IfNoValueTF{#1}{\operatorname{abandonment}}{\operatorname{abandonment@}\!#1}
}
\newcommand{\coverage}{\operatorname{cov}}

\NewDocumentCommand{\fmacro}{o}{
    \operatorname{F}^{\text{macro}}\IfNoValueTF{#1}{}{_{#1}}
}
\NewDocumentCommand{\topk}{o}{
    \IfNoValueTF{#1}{\operatorname{top}}{\operatorname{top}_#1}
}
\NewDocumentCommand{\rankof}{m}{
    \operatorname{r}_#1
}

\newcommand{\phijain}{\phi_{\xmlcpropensitymodel}}
\newcommand{\phijpv}{\phi_{\xmlcpropensitymodel}}
\newcommand{\phipower}{\phi_{\text{P}}}
\newcommand{\phirichards}{\phi_{\text{R}}}

%% file: 1-intro.tex
Extreme multi-label classification (XMLC) is a supervised learning problem 
where only a few labels from an enormous label space, 
reaching orders of millions, are relevant per data point. 
Notable examples are 
tagging of text documents~\citep{Agrawal_et_al_2013},
content annotation for multimedia search~\citep{Deng_et_al_2011}, 
and diverse types of recommendation, 
including webpages-to-ads~\citep{Beygelzimer_et_al_2009b}, 
ads-to-bid-words~\citep{Prabhu_Varma_2014}, 
users-to-items~\citep{Zhuo_et_al_2020}, 
queries-to-items~\citep{Medini_et_al_2019}, 
or items-to-queries~\citep{Chang_et_al_2020}.
These practical applications pose statistical challenges, including: 
1) long-tailed distribution of labels---infrequent (tail) labels are much harder to predict 
than frequent (head) labels due to data imbalance, and a model completely ignoring the tail labels can get very high scores on standard performance metrics;  
2) missing relevant labels in the observed training data---since it is nearly impossible to check the whole set of labels 
when it is so large.

To address the latter issue, 
propensity-scored versions of popular measures (i.e., precision$@k$ and nDCG$@k$) 
were introduced by~\citet{Jain_et_al_2016}. 
Under the propensity model, it is assumed that an assignment of a label to an example is always correct,
but the supervision may skip some positive labels, and propensity of a label refers to the probability of not skipping that label.
Under the implicit assumption that the chance for a label to be missing is higher for tail than for head labels, the propensity-scored measures were used to evaluate the prediction performance on tail labels. 
Despite being originally introduced to study the phenomenon of missing labels in XMLC, over the years, they have found their way into the literature as default performance metrics on tail labels \citep{You_et_al_2019,guo_breaking_2019,Babbar_Scholkopf_2019}.

In this work, we take a step back and thoroughly investigate the validity of the propensity model of~\citet{Jain_et_al_2016}, 
further referred to as \xmlcpropensitymodel{} (from the first letters of authors' names),
for the dual usage of missing and long-tail labels in XMLC.
We start our discussion by recalling the definition of the XMLC problem, stating the problem of missing labels, 
and bringing closer the issues with long-tail labels (Section~\ref{sec:problem_statement}).
We recall the \xmlcpropensitymodel{} propensity model and highlight its shortcomings in Section~\ref{sec:critical-view-on-propensities}, both in terms of the model itself and in regard to its current usage in XMLC.
In particular, we demonstrate that this model: 
(i) does not fulfill natural conditions that may be desired of a reasonable propensity model, 
(ii) falls short on reliable and reproducible estimation of the model hyper-parameters,
and (iii) leads to implausible results exceeding substantially the natural range of the metrics
(e.g., precision@k $>$ 300\%).

After formally studying the above short-comings of the \xmlcpropensitymodel{} model, 
we propose a suite of alternatives (c.f. Section \ref{sec:recipes-to-follow}) 
which are promising to follow for a more principled approach in 
(i) evaluating machine learning systems trained on data with incomplete user feedback, 
and (ii) disentangling the individual contribution of missing and tail labels in XMLC. 
We suggest using unbiased sets for validating models designed to deal with missing labels.
Alternatively, one should use a set with a controlled bias, 
on which one can obtain unbiased estimates of performance metrics.
Thereafter, we discuss alternative propensity models, which possess desirable analytical properties, 
and compare them with the \xmlcpropensitymodel{} model empirically, confirming its shortcomings.
We also show the efficacy of a framework 
in which label propensities and parameters of the learning model are learned jointly. 
Towards disambiguating the phenomenon of missing and long-tail labels in XMLC, 
we finally highlight other metrics as possible options for measuring tail-label performance instead of conflating these with missing labels.

It should be noted that, unlike most contemporary advances in XMLC, our goal in this work is not algorithmic. 
Instead, we take a critical viewpoint and study the commonly-used propensity model, 
explicating the consequences when it is used in real-world production environments.

%% file: 2-xmlc.tex
The goal of XMLC is to find a mapping between instances $\rinstance \in \instancespace$ 
and a finite set of $\numlabels$ non-mutually-exclusive class labels.%
\footnote{We use capital letters for random variables, 
and calligraphic letters for sets.} 
This means that any specific
realization $\sinstance$ of $\rinstance$ is associated with a (possibly empty)
subset $\relevantlabels{\sinstance} \subset [\numlabels]$ of the labels called
the \emph{relevant} or \emph{positive} labels, 
with the complement, $[\numlabels] \setminus \relevantlabels{\sinstance}$, 
of the \emph{irrelevant} or \emph{negative} ones. 
We identify the relevant labels with a binary vector
$\slabels \in \labelspace$ through $\sblabel_j = \indicator[j \in
\relevantlabels{\sinstance}]$,\footnote{$\indicator[\cdot]$ is the indicator function.}
where $\labelspace \coloneqq \set{0,
1}^\numlabels$ is called the \emph{label vector space}. In the classical setting, we
assume that observations $(\rinstance, \rlabels)$ are generated independently
and identically according to a probability distribution $\probability$ on
$\instancespace \times \labelspace$.  
In case of XMLC we assume $\numlabels$ to be a large number (e.g., $\ge 10^5$), 
and $\|\rlabels\|_1$ to be much smaller than $\numlabels$, $\|\rlabels\|_1 \ll \numlabels$.

The problem of XMLC can be defined as finding a \emph{classifier} 
$\defmap{\hypothesis}{\instancespace}{\reals^m}$ which minimizes the \emph{task risk}:
\begin{equation}
\risk_{\taskloss}[\hypothesis; \rinstance, \rlabels] \coloneqq \expectation[\taskloss(\rlabels, \hypothesis(\rinstance))]\,,
\end{equation}
where $\defmap{\taskloss}{\labelspace \times \reals^m}{\nnreals}$ is the (\emph{task}) \emph{loss}.
The optimal (\emph{Bayes}) classifier for a loss $\taskloss$ is given by
\begin{equation}
\bayescls(\sinstance) = \argmin_{\spreds \in \reals^m} \expectation[\taskloss(\rlabels, \spreds) \mid \rinstance = \sinstance]\,.
\end{equation}
The above definitions follow the standard statistical learning framework. 
Let us notice, however, that in XMLC, instead of loss functions, 
one often uses performance metrics,
which are rather maximized than minimized.
Moreover, these definitions correspond to the most natural setting
in which a decision is made based on a single $\sinstance$. 
Later in the paper, we also consider more general metrics 
that cannot be optimized with respect to individual instances.

Typically, a task loss is hard to optimize and one chooses
instead a \emph{surrogate loss} that is easier to cope with, e.g., because it is
differentiable and convex. Furthermore, instead of a
probability distribution, a learning algorithm operates on a finite i.i.d.
sample and minimizes the corresponding empirical risk.

\subsection{Missing labels}
\label{subsec:missing_labels}
In XMLC, the observed data might not follow the distribution we want to learn about. 
As an illustrative example, take the Wikipedia-500k dataset. 
The content of a Wikipedia article should be matched with a set of categories the article belongs to. 
Such a dataset can be easily created by scraping existing Wikipedia annotations. 
However, there are about \num{500000} categories on Wikipedia, 
and it is clear that the original authors and curators have
never checked every single category for each article.%
\footnote{If it took a human one second to check a category for an article, 
then annotating a single article fully would take almost 6 days.}
On the other hand, each category that has been assigned to an article has been
verified by a human to be relevant. 
Therefore, the labeling error can be assumed to be strongly one-sided: 
There may be many missing labels, but spurious labels should be uncommon. 

To contrast ground-truth labels $\rlabels$ with those actually available, 
we denote the \emph{observed labels} $\rolabels$ using a tilde.%
\footnote{Note that other papers, including~\cite{Jain_et_al_2016}, 
use often a slightly different notation.} 
Mathematically, the setting studied in this paper is defined by
\begin{equation}
 \probability[\rolabels \preceq \rlabels \given \rinstance] = 1\,, \qquad
 \probability[\rolabels \npreceq \rlabels \given \rinstance] = 0\,,
\label{eqn:general_missing_labels_problem}
\end{equation}
where $\rolabels \preceq \rlabels$ means that $\rolabel_j \le \rblabel_j$ for all $j \in [m]$,
and $\rolabels \npreceq \rlabels$ means that 
there is at least one label for which $\rolabel_j > \rblabel_j$. 
Notice that the above equations cover also the no noise case, 
as we may have $\probability[\rolabels = \rlabels \given \rinstance] = 1$. 

Reconstruction of the ground truth distribution from the observed one, in the general case,
is not a trivial task from the statistical and computational perspective, 
as it requires an exponential number of parameters.
Let $\eta_{\slabels}(\sinstance) \coloneqq \probability[\rlabels=\slabels \given \rinstance = \sinstance]$
and $\tilde{\eta}_{\slabels}(\sinstance) \coloneqq \probability[\rolabels=\slabels \given \rinstance = \sinstance]$.
We have then
\begin{equation}
\tilde{\eta}_{\solabels}(\bx) = 
\sum_{\slabels} 
p_{\solabels}(\slabels, \sinstance)
\eta_{\slabels}(\bx)\,,
\end{equation}
where $p_{\solabels}(\slabels, \sinstance) \coloneqq \probability[\rolabels=\solabels \given \rlabels=\slabels, \rinstance = \sinstance]$ is 
a propensity of observing $\solabels$ for ground-truth labels $\slabels$ and instance $\sinstance$.
Notice that from (\ref{eqn:general_missing_labels_problem}) 
we have $p_{\solabels}(\slabels, \sinstance) = 0$ for $\solabels \npreceq \slabels$.
Furthermore, let $\vec{\eta}_{\calY}(\sinstance)$ and $\tilde{\vec{\eta}}_{\calY}(\sinstance)$
be vectors of $\eta_{\slabels}(\sinstance)$ and $\tilde{\eta}_{\slabels}(\sinstance)$, respectively,
for all $\slabels \in \labelspace$ given in some predefined order $\pi$. 
Let $\mathsf{C}$ be a matrix containing all propensities $p_{\solabels}(\slabels, \sinstance)$,
with rows and columns corresponding to $\rolabels$ and $\rlabels$, respectively, 
and organized according to $\pi$.  
Then, we get: 
\begin{equation}
\tilde{\vec{\eta}}_{\calY}(\sinstance) = \mathsf{C} \vec{\eta}_{\calY}(\sinstance) \,,
\end{equation}
and, finally:
\begin{equation}
\vec{\eta}_{\calY}(\sinstance) = \mathsf{C}^{-1}\tilde{\vec{\eta}}_{\calY}(\sinstance) \,,
\end{equation}
where we need to assume that $\mathsf{C}$ is invertible.

Because of the practical reasons, 
a much simpler, label-wise, propensities  
are commonly used that are defined for each label separately:
\begin{equation}
 \propensity_j(\rinstance) \coloneqq \probability[\rolabel_j=1 \given \rblabel_j = 1, \rinstance]\,.
\label{eq:missing-model}
\end{equation}

Let $\tilde{\eta}_{j}(\sinstance) \coloneqq \probability[\rolabel_j = 1 \given \rinstance = \sinstance]$ 
and $\eta_{j}(\sinstance) \coloneqq \probability[\rblabel_j = 1 \given \rinstance = \sinstance]$.
We have then:
\begin{equation}
\olabeldist_j(\sinstance) = \propensity_j(\sinstance) \labeldist_j(\sinstance)\,, \qquad
\labeldist_j(\sinstance) = \olabeldist_j(\sinstance) / \propensity_j(\sinstance)\,.
\label{eqn:prob_adjustment}
\end{equation}

If propensities are known, then they can be used to construct an unbiased, task or 
surrogate, loss $\ubloss$ \citep{van_rooyen_theory_2017}
in the sense that
\begin{equation}
    \forall \hypothesis: \; \risk_{\lossfn}[\hypothesis; \rinstance, \rlabels] = \risk_{\ubloss}[\hypothesis; \rinstance, \rolabels].
    \label{eq:unbiased-risk-property}
\end{equation}
The construction of the unbiased counterpart depends on the 
form of propensities,  
e.g., the label-wise propensities (\ref{eq:missing-model}) are sufficient
for losses decomposable over labels~\citep{natarajan_cost-sensitive_2017}  
like Hamming loss or binary cross-entropy,
but might not be for more complex losses without additional assumptions~\citep{Schultheis_Babbar_2021}.
The unbiased losses can be used in training procedures~\citep{Jain_et_al_2016,Qaraei_et_al_2021}
or for estimating the performance of classifiers.
For some losses, such as Hamming loss or precision@$k$, the Bayes
classifier can be written as a function of the conditional label distributions
$\labeldist_j(\sinstance)$.
In this case, one can adjust existing inference procedures to use (\ref{eqn:prob_adjustment}) 
to obtain estimates of $\labeldist_j(\sinstance)$ 
from estimates of $\olabeldist_j(\sinstance)$~\citep{propensity-plt}.

\subsection{Long-tailed label distribution}
\label{subsec:long-tail-labels}

A defining characteristic of extreme classification data is that the label
distribution is highly imbalanced. In the binary case, the amount of imbalance
is completely determined by the imbalance ratio
$\frac{\probability[\rblabel=0]}{\probability[\rblabel=1]}$. 
In this sense, almost every binary problem
corresponding to a label is highly imbalanced in XMLC, i.e., only a small
fraction of training instances will be associated with that label. However,
in XMLC, the data are also imbalanced when comparing different labels. In
analogy to the binary case, we can define an inter-label imbalance ratio
through $\operatorname{ILIR} = \frac{\max\{\probability[\rblabel_i=1] : i \in [\numlabels]\}}{\min\{ \probability[\rblabel_j = 1] : j \in
[\numlabels] \}}$.\footnote{This concept is also used in \emph{multiclass}
classification, e.g. the \emph{imbalance factor} of \citet{cui2019class}.}
Nevertheless, the imbalance factor does not cover an important
property of the label distribution. It could be that most labels have a
large number of positives, but some have very few, or vice-versa. The latter
case is what happens in XMLC, where the label distribution
is said to be \emph{long-tailed}~\citep{Bhatia_et_al_2015, Babbar_Scholkopf_2017}.

\begin{table}
\small
\caption{Imbalance characteristics of typical XMLC datasets.}
\label{table:dataset-stats}
\begin{filecontents*}{data/datasets.txt}
{dataset}           {instances}     {head imbalance}        {class imbalance}       {obesity}       {80percent}
Eurlex-4K           15539           15                      1006                    85              19.9 
AmazonCat-13K       1186239         3.3                     355211                  90              4.8
Wiki10-31K          14146           1.2                     11411                   72              19.6
Delicious-200K      196606          3                       64548                   85              4.0
WikiLSHTC-325K      1778351         6.1                     293936                  84              20.7
Wikipedia-500K      1813391         6.5                     279506                  80              25.1
Amazon-670K         490449          268                     1826                    66              54.3
Amazon-3M           1717899         143                     12014                   83              26.1
\end{filecontents*}
\pgfplotstabletypeset[
every head row/.style={
  before row={\toprule},
  after row={\midrule}},
  columns={dataset,instances,head imbalance,class imbalance,80percent},
every last row/.style={after row=\bottomrule},
columns/dataset/.style={string type, column type={l|}, column name={Dataset}},
columns/head imbalance/.style={column type = {r}, column name={min IR},zerofill,precision=1},
columns/instances/.style={column type = {r}, column name={Instances}, precision=2, sci,zerofill},
columns/class imbalance/.style={column type = {r}, column name={ILIR}, sci,zerofill, precision=2},
columns/80percent/.style={column type = {r}, column name={Pos-80\%},zerofill,precision=1}
]{data/datasets.txt}
\end{table}

In \autoref{table:dataset-stats}, these imbalance measures are shown for
several XMLC datasets. 
We use \textit{min IR} to denote the binary imbalance ratio of the head label,
i.e., the label with the largest fraction of positive instances
(therefore, its IR is the smallest),
and Pos-$80\%$ to indicate the minimum fraction of class labels that retain $80\%$ of 
positive labels (i.e., $y_j = 1$) in the training set.
For example, in Delicious-200K, only four percent of the class labels contain $80\%$ of the positive labels.

In addition to the number of positive instances of a sparse label, 
their distribution within the feature space can be very important. 
If $\rblabel_j$ has few positives, but in a small pocket $\instancespace_{+}$ of
the feature space it still fulfills 
$\probability[\rblabel_j = 1 \mid \rinstance \in \instancespace_{+}]\approx 1$, 
then a learning algorithm might still learn a reasonable decision boundary, 
especially if the overall number of samples is large enough \citep[p. 23]{fernandez2018learning}. 
In contrast, if $\probability[\rblabel_j = 1 \mid \rinstance] \ll 1$ 
everywhere (called \emph{uniform class imbalance} by \citet{singh2021statistical}), 
learning to recognize the given class might be infeasible.

%% file: 3-critical-view.tex
The goal of \citet{Jain_et_al_2016} was to develop loss functions for XMLC that
\begingroup
\addtolength\leftmargini{-0.1in}
\begin{quote}
(a) prioritize predicting the few relevant labels over the large number of
irrelevant ones; (b) do not erroneously treat missing labels as irrelevant
[...] (c) promote the accurate prediction of infrequently occurring, hard to
predict, but rewarding tail labels.
\end{quote}
\endgroup

There are two main contributions of the paper that are relevant for our discussion:
First, the development of unbiased loss functions that allow compensating for missing
labels if their propensities are known, and second, an empirical model to estimate
these propensities on XMLC data.

\subsubsection{Propensity-scored losses}

Popular XMLC performance metrics
focus on the highest scored labels by the prediction algorithm.  
Examples of such metrics are \emph{precision at k} ($\precision[k]$), \emph{recall at k} ($\recall[k]$), 
or \emph{(normalized) discounted cumulative gain} $\ndcg[k]$.
For these metrics, unbiased estimates 
in the sense of \eqref{eq:unbiased-risk-property} can be calculated, 
which are called the \emph{propensity-scored} (PS) variants of these metrics
(more examples in \citep[Table 1]{Jain_et_al_2016}).
Table~\ref{tab:psmetrics} gives the formal definitions,
where $\topk[k]$ maps a vector to the indices of its top-k components, 
$\rankof{j}(\spreds)$ gives the ranking of the $j$-th element in the vector,
and $p_j$ is the propensity for label $j$. 
Let us notice that, in general, $p_j$ shall depend on $x$, 
but \citet{Jain_et_al_2016} practically assume $p_j$ to be a constant value for each label $j$.
Moreover, of the above unbiased estimates, only $\psprecision$ and $\psndcg$ have found widespread use
\citep{Bhatia_et_al_2016}, because $\psrecall$ still requires the knowledge of
the total number of relevant labels $\|\slabels\|_1$.
\input{tables/xmlc-metrics}

Because \citet{Jain_et_al_2016} observed that the unbiased estimates
could results in values larger than one, they suggest 
a normalized version of these metrics to be reported (cf. Section~\ref{subsec:implausible_results}).
In subsequent literature, the distinction between the unbiased metrics
and the normalized versions is not always preserved, e.g., \citet{Bhatia_et_al_2016} present unbiased formulas
but lists normalized values.

\subsubsection{Empirical propensity model}
\label{subsec:empirical-model}
In order to use the propensity-scored loss functions, one needs to have
the propensities available for the individual labels. Since true propensities
are unknown for the XMLC benchmark datasets,
\citet{Jain_et_al_2016} proposed to model propensities
as a function of labels frequencies, 
resulting in propensities being a constant value for each label. 

Let $\phi$ denote a propensity model.  
The model defined in~\citep{Jain_et_al_2016}
can be expressed via label priors $\olabelprior_j \coloneqq \probability[\rolabel_j=1]$:
\begin{equation}
\label{eq:jain-prop-model}
\!\!\! \propensity_j \!=\! \phijain(\olabelprior_j; \numinstances, a, b) \!\coloneqq\! \frac{1}{1 + (\log \numinstances - 1)(b + 1)^a e^{-a \log (\numinstances \olabelprior_j + b)}} \,,
\end{equation}
where $n$ is the number of training instances, and $a$ and $b$ are
dataset-dependent parameters.

In order to arrive at this model and determine values for $a$ and $b$,
\citet{Jain_et_al_2016} investigated two datasets in which ancillary
information could be used to identify some missing labels.

For a Wikipedia-based dataset, the parameters of the model have been
estimated with the help of a label hierarchy. They assumed
that if a label is relevant to an article, then all its ancestors in the
hierarchy should also be relevant. If not present, they are counted as
missing. This allows plotting the fraction of instances in which the label is
missing over the number of instances in which it appears. This seems to
follow a sigmoidal shape as described by \eqref{eq:jain-prop-model}, see~Figure~\ref{fig:wikipedia-ps-est}. 
The
parameters $a$ and $b$ were then determined by fitting the model against the
estimated values, where only labels with more than 4 descendants were used to
improve robustness. The obtained values are $a = 0.5, b = 0.4$.
\input{figures/wikipedia-ps-est}

For the Amazon data set, which is an item-to-item recommendation task, 
missing labels have been approximated using ``also viewed'' and ``also bought`` information. 
It was assumed that a label $j$ (an item) is relevant to all the items viewed 
along with items that were also bought with label $j$, as proposed by \citet{McAuley_et_al_2015}. 
The obtained values are $a = 0.6$ and $b = 2.6$.

For other data sets the authors propose, if there is no other possibility of
estimating parameters $a$ and $b$, to use averages of the values obtained for
Wikipedia and Amazon data sets (which are $a = 0.55$, $b = 1.5$). This, in fact,
has become a standard followed in many papers without questioning its rationality.

The above propensity model is then typically used in the metric of choice 
for model selection and evaluation. 
It has also been incorporated into training procedures. 
For example, decision tree methods can directly use the
propensity-scored variants of metrics such as precision$@k$ or nDCG$@k$ \citep{Jain_et_al_2016}.
Alternatively, one can use unbiased or upper-bounded propensity-scored
surrogate losses~\citep{Qaraei_et_al_2021}.

\subsubsection{Propensity and long tails}
\label{subsec:prop-long-tail}
The form of \eqref{eq:jain-prop-model} implies that tail labels are assigned
lower propensities, which means that in metrics like those in \autoref{tab:psmetrics} these
tail labels, if predicted correctly, will be weighted more strongly than head
labels. In particular, the resulting weightings resemble existing weighting
schemes used for long-tailed learning tasks, leading the authors to conclude:

\begingroup
\addtolength\leftmargini{-0.1in}
\begin{quote}
Such weights arise naturally as inverse propensities in the unbiased losses
developed in this paper. [...] This not only provides a sound theoretical
justification of label weighting heuristics for recommending rare items but
also leads to a more principled setting of the weights.
\end{quote}
\endgroup

As a result, propensity-scored variants are also viewed as metrics in their own right, and are currently used both 
to counteract missing labels (as unbiased estimates) and to weigh tail labels (as independent metrics),
becoming established performance metrics commonly used in XMLC.%
\footnote{We list several examples of references to propensity-scored losses:
``We examined the performance on tail labels by PSP@k''~\citep{You_et_al_2019};
``We achieve high precision and propensity scores, 
thus demonstrating the effectiveness of our method even on infrequent tail labels.''~\citep{guo_breaking_2019};
``capture prediction accuracy of a learning algorithm at top-k slots of prediction, and also the diversity of
prediction by giving higher score for predicting rarely occurring
tail-labels''~\citep{Babbar_Scholkopf_2019};
``propensity scored precision@k which has recently been shown to be an
unbiased, and more suitable, metric''~\citep{jain_slice_2019};
``which leads to better performance on tail labels.''~\citep{Yen_et_al_2017};
``propensity scored variant which is unbiased and assigns higher rewards for accurate tail label predictions'', 
``evaluate prediction performance on tail labels using propensity scored variants''~\citep{Khandagale_et_al_2019};
``replacing the nDCG loss with its propensity scored variant and using additional classifiers designed for tail labels''~\citep{Tagami_2017}.
}

\subsection{Discussion of missing-labels assumptions}

In order to derive unbiased loss functions, we need to impose assumptions on
the process of how labels go missing, as initially discussed in Section~\ref{subsec:missing_labels}.
Unfortunately, \citet{Jain_et_al_2016} sent a potentially misleading message in this regard. 
Their Theorem 4.1 proves
\begin{equation}
    \expectation[\lossfn(\rlabels, \spreds)] = \expectation[\ubloss(\rolabels, \spreds)]\,,
\end{equation}
for any \textbf{fixed} prediction $\spreds$ without a clear dependence on $X$. 
This also implies that the assumptions behind the propensities are unclear.
Even if we assume the propensities to be constant for label $j$,
the exact form of this assumption is necessary to properly prove unbiasedness 
in the sense of (\ref{eq:unbiased-risk-property}).
Notice that $\probability[\rolabel_j = 1 \given \rblabel_j = 1] = p_j$
does not imply $\probability[\rolabel_j = 1 \given \rblabel_j = 1, X] = p_j$.
Moreover, for more complex functions, such as recall$@k$, this assumption may take the form of 
$\probability[\rolabel_j = 1 \given \rblabel_j = 1, \rlabels_{\!\neg j}, X] = p_j$,
where  $\rlabels_{\!\neg j}$ represents ground-truth labels without label $j$
(see Appendix for an example).

In general, we cannot expect the independence of missing labels from the
instance's features to hold. Consider, for example, cases where the feature and
label space are of a similar origin~\citep{dahiya2021siamesexml}, such as matching
Wikipedia titles or articles to categories. 
It seems unlikely that a label such as ``Italy'' would be missing
for articles containing the word ``Italy'' in the subject, 
but it might be missing for articles that still pertain to Italy  
but do not feature the word ``Italy'' in the title.
The assumption that propensities are constant for each label simplifies the model 
significantly and leads to much simpler computational procedures. 
Unfortunately, if this assumption is not satisfied, 
then one may get implausible results as discussed later. 

The assumption that the propensities do not depend on other labels going missing 
does not need to hold in practice as well.
For example, a user that tagged the article for ``Italy'' with ``Member states
of the European Union'' might be primed to think of more examples of
organizations in which Italy is a member, and thus e.g., ``Current member
states of the United Nations'' might be less likely to be forgotten than if
the EU membership had been forgotten. 
Fortunately, in many cases, the unbiased
estimate does not actually require this dependence --- 
if the loss function can be written as a sum over contributions from each label individually, 
then the labels do not interact with each other 
and the label-wise properties are sufficient to obtain unbiased losses.
This is the case for the popular $\psprecision$ and $\psndcg$ metrics.

\subsection{Shortcomings of the \xmlcpropensitymodel{} propensity model}
\label{subsec:shortcomings}

Let us discuss several issues of the \xmlcpropensitymodel{} model,
concerning theoretical and empirical shortcomings, 
as well as some problems in the way the parameters of the model have been established.

\paragraph{Scaling behavior}
Let us first observe that \eqref{eq:jain-prop-model} does not preserve propensity estimates 
if the amount of data is changed, without changing its characteristics, 
e.g., by sub- or over-sampling the dataset. 
In particular, if one increases the amount of available data 
by making multiple copies of the dataset, 
which should not change the estimates of label priors $\tilde \pi_j$
given by $\tilde{\numinstances}_j/\numinstances$ 
(with  $\tilde \numinstances_j$ being the number of positive instances of label $j$ 
in the observed, noisy training set), 
the \xmlcpropensitymodel{} model will estimate propensities to be equal one, 
i.e., no missing labels, as the amount of data goes to infinity:
\begin{multline}
    \!\!\!\!\!\lim_{\numinstances \rightarrow \infty} \phijain(\tilde \pi_j,  n) = \frac{1}{1 + (b + 1)^a  \lim_{\numinstances \rightarrow \infty} (\log \numinstances - 1)e^{-a \log (\tilde\pi_j \numinstances + b)}} \\
    = \frac{1}{1 + (b + 1)^a \lim_{\numinstances \rightarrow \infty} (\log \numinstances) (\tilde\pi_j \numinstances)^{-a}} = 1.
\end{multline}
This means that we cannot interpret $a$ and $b$ as parameters of some
underlying (unknown) process that describes the labeling process. As we cannot
even have fixed $a$ and $b$ when the data come from the same process, this
very much calls into question the approach of using values for $a$ and $b$
across datasets as is current practice.

\paragraph{Estimation process}
Setting aside structural concerns about \eqref{eq:jain-prop-model}, the
estimation of the parameters $a$ and $b$ still remains an issue. First, by
identifying missing labels based on meta-data as described in
Section~\ref{subsec:empirical-model}, only an upper-bound on the propensity is
estimated, since labels may also be missing in other ways. 
For example, we tried to reproduce the procedure of propensity estimation on the Wikipedia dataset. 
We have found that only around \num{40000} out of \num{500000}
labels meet the criteria of the sufficient number of descendants selected by the authors,
and around \num{300000} labels are without descendants, 
so they would never be considered missing by this protocol. 

Further, one might argue that in cases in which missing labels can be
identified by some side-channel information such as label hierarchies, then
one can directly impute these missing labels and need not worry about training
with missing labels. 

\paragraph{Propensity as a function of frequency}
This still leaves the question of whether such estimates are sensible.
Even though there is clearly a trend that labels within a given range of
frequency have -- on average -- a certain propensity, for each individual label
the actual propensity can fluctuate widely around this mean, as shown in 
\autoref{fig:wikipedia-ps-est} that we obtained following the original procedure for estimating propensities.

\paragraph{Reproducibility} 

The description of the process of propensity estimation in~\citep{Jain_et_al_2016} 
is rather sparse on details. 
While meta-data for Wikipedia is easily obtainable, 
it is not clear what is the source of ancillary information 
that has been used for the Amazon dataset. 
Additionally, depending on the preprocessing steps 
and criteria, such as the number of descendants in the label hierarchy, 
one can achieve very different estimates of parameters $a$ and $b$.

\subsection{Implausible results and normalization}
\label{subsec:implausible_results}

Despite being unable to verify the correctness 
of the assumptions and the \xmlcpropensitymodel{} model 
without actual clean ground truth data, 
we are still able to show 
that the approach of~\citet{Jain_et_al_2016} leads to implausible results.
For example, $\psprecision[k]$, 
as an unbiased estimate of $\precision[k]$ on the ground-truth data,
should be bounded between zero and one.
However, when calculating this measure for a real classifier, 
the result may exceed this range substantially. 
Of course, for an individual instance or a small subset of them, 
the unbiased estimate does not need to fall into that range, 
but a large deviation from the true value becomes exceedingly unlikely 
when averaging over the entire dataset. 

To circumvent this issue, \citet{Jain_et_al_2016} suggest 
to report a normalized version of $\psprecision[k]$, 
also calling this measure ``propensity-scored precision''.
The normalization is realized by dividing the metrics value 
by the largest possible value that any prediction could have achieved on that data:
\begin{equation}
\label{eq:ps-loss-norm}
\textrm{Norm}\psprecision[k] = \frac{\sum_{i=1}^{n} \psprecision[k](\solabels_i, \spreds_i)}{\sum_{i=1}^{n} \max_{\mathbf{z}} \psprecision[k](\solabels_i, \mathbf{z})} \,.
\end{equation}
The normalization introduces a factor 
that is constant over the entire dataset, 
and thus does not influence model selection. 
However, it removes the interpretation of the received value 
as an unbiased estimate of the metric on clean data, 
and it hides the model misspecification.
Table~\ref{tab:norm-unnorm-psp-at-k} reports the values of both variants of $\psprecision[k]$,
showing how severe this issue is. 

\input{tables/norm-unnorm-psp-at-k}

\subsection{The current use of propensity metrics}
\label{subseq:usage-problems}

It seems that the current use of propensity metrics mixes up, 
in a not entirely clear way, two different issues, missing and tail labels.
As mentioned in Section~\ref{subsec:prop-long-tail},
these metrics might be used for the purpose of giving 
more weight to tail labels. 
In this case, the normalization step seems to be a valid procedure. 
However, a propensity metric loses its original interpretation, 
and it is just one way of accounting for tail labels, 
without any concrete justification. 
For this use case, it would be preferable to have a metric 
that treats tail labels in a principled way. 
As a first step towards that goal, 
Section~\ref{sec:task-losses-for-long-tails} provides 
some discussion on alternative task losses.

Only in the interpretation as a tail-performance promoting loss, 
it does make sense to speak of a trade-off in performance 
between vanilla and propensity-scored metrics, as these are conceptually different. 
In the missing-labels interpretation, taking the propensities into account is not
calculating a different conceptual metric, but instead, the \emph{correct} way
of calculating the unweighted, but the true performance of a classifier.
Of course, in XMLC, both interpretations can be combined, 
i.e., one would like to have a task loss that is adapted to tail labels, 
but calculate it in a way that takes missing labels into account. 
The closest to this in the literature is \citep{Qaraei_et_al_2021}, 
where training uses a loss that combines unbiased estimates and class-rebalancing, 
but still, evaluation is performed using vanilla and propensity-scored metrics, 
instead of a propensity-scored variant of a tail-weighted metric.

%% file: tables/xmlc-metrics.tex
\begin{table}[h!]
    \caption{Definitions of popular XMLC performance metrics and their unbiased estimates on missing labels.}
    \label{tab:psmetrics}
    \begin{center}
    \resizebox{\linewidth}{!}{
        \begin{tabular}{l|c|c}
             \toprule
             Measure & Definition & Unbiased estimate \\
             \midrule
             $\precision[k](\slabels, \spreds)$  & 
             $k^{-1} \sum_{j \in \topk[k](\spreds)} \sblabel_j$ 
             & $k^{-1} \sum_{j \in \topk[k](\spreds)} \solabel_j / \propensity_j, \label{eq:def:psp}$ \\
             \midrule
             $\recall[k](\slabels, \spreds) $ 
             & $\|\slabels\|_1^{-1}\sum_{{j \in \topk[k](\spreds)}} \sblabel_j$ 
             & $\|\slabels\|_1^{-1}\sum_{j \in \topk[k](\spreds)} \solabel_j/ \propensity_j$ \\
             \midrule
             $\ndcg[k] (\slabels, \spreds) $ 
             & $\frac{\sum_{{j \in \topk[k](\spreds)}} \frac{\sblabel_j}{\log(\rankof{j}(\spreds)+1)}}{\sum_{j=1}^{k} \frac{1}{\log(j+1)}}$
             & $\frac{\sum_{j \in \topk[k](\spreds)} \frac{\solabel_j}{\propensity_j \log(\rankof{j}(\spreds)+1)}}{
    \sum_{j=1}^{k} \frac{1}{\log(j\!+\!1)}}$\\
             \bottomrule
        \end{tabular}
    }\end{center}
\end{table}

%% file: figures/wikipedia-ps-est.tex
\begin{figure}[t]
\centering
\includegraphics[width=0.75\linewidth]{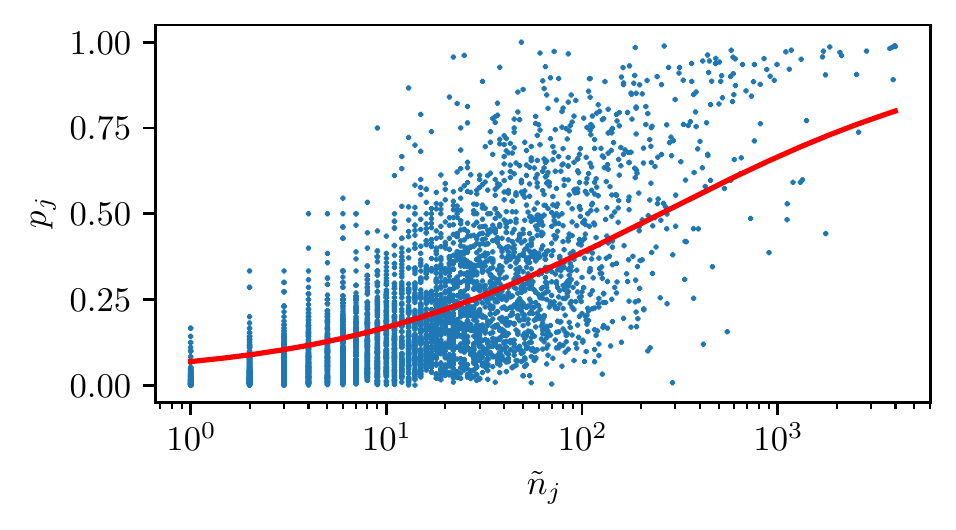}
\vspace{-0.4cm}
\caption{
Reproduced estimates of propensities for Wikipedia-500K dataset using labels hierarchy and propensity function $\phi_{\xmlcpropensitymodel}$ with $a=0.5$ and $b=0.4$ as estimated by \citet{Jain_et_al_2016} for this dataset.}
\label{fig:wikipedia-ps-est}
\end{figure}

%% file: tables/norm-unnorm-psp-at-k.tex
\begin{table}[h]
    \caption{Normalized and unnormalized propensity-scored precision of PfastreXML~\citep{Jain_et_al_2016}, when using the \xmlcpropensitymodel{} model, with $a=0.5$, $b=0.4$ for WikiLSHTC-325K and $a=0.6$, $b=2.6$ for Amazon-670K.} 
    \label{tab:norm-unnorm-psp-at-k}

    \centering
    \small
    \begin{tabular}{c|rrr|rrr}
        \toprule
        & \multicolumn{3}{c|}{WikiLSTHC-325K} & \multicolumn{3}{c}{Amazon-670K} \\
        \midrule
        PSP(\%) & $@1$ & $@3$ & $@5$ & $@1$ & $@3$ & $@5$ \\
        \midrule
             Normalized & 31.16 & 31.80 & 33.35 & 29.93 & 31.26 & 32.80 \\
             Unnormalized & 196.96 & 118.54 & 85.28 & 326.47 & 282.28 & 250.57 \\
        \bottomrule
    \end{tabular}
\end{table}

%% file: 4-recipes.tex
In this section, we present several recipes on how to conduct 
research on missing and tail labels in XMLC.
We start our discussion with a recommendation of using an additional dataset
which is either unbiased or its bias is under control.
We then discuss several alternatives for the \xmlcpropensitymodel{} model,
show how to fit the models to unbiased data, and how to compare them empirically.
Next, we introduce methods that jointly train the prediction and the propensity model. 
Despite our critical remarks, we consider in this section only propensities being constant for each label.
Finally, we discuss performance metrics for long-tail labels 
that might be a better choice than propensity-scored metrics.%
\footnote{Repository with the code to reproduce all experiments: \url{https://github.com/mwydmuch/missing-labels-long-tails-and-propensities-in-xmlc}}

\subsection{Bias-controlled validation and test sets}
\label{sec:unbiased-sets}

If indeed our training data are biased by missing labels, 
the best way to test, validate, and estimate the propensity model 
is to use unbiased data or data with controlled bias.
In the latter case, the bias is controlled in such a way that 
unbiased estimates can be easily computed.
Such data are definitely costly to get, but even a small set can be beneficial,
helping in selecting the right model to be used in production and in estimating its real performance. 
This is a standard approach used in recommendation systems~\citep{Saito_et_al_2020,Yang_et_al_2018} 
and search engines~\citep{Joachims_et_al_2018}. 
Even if, in a given application, it is not possible to obtain an unbiased data due to some constraints,
we should investigate algorithms on benchmarks with unbiased or bias-controlled sets associated. 
Without this investigation, it is hard to verify which methods are indeed working and which are misleading. 
In many real-world applications, the final evaluation of the model 
is performed in A/B tests on real, unbiased data. 
We should avoid situations in which the results of A/B tests are against 
our expectations coming from offline experiments.

To follow the above recommendation, 
we have prepared several datasets of different types,
which can be used for experimentation with missing labels in XMLC. 
The first type contains fully synthetic datasets. 
The second one is a modification of the standard XMLC benchmarks. 
The final dataset is a variant of the Yahoo R3 dataset,\footnote{\url{https://webscope.sandbox.yahoo.com/}}
transformed from a recommendation problem to multi-label classification. 
Statistics of these datasets along with additional information are given in Appendix.

The synthetic datasets are generated in a similar way as in~\citep{Jimena_et_al_2014_ml_datagen}.
They are parameterized by the number $m$ of labels. 
In the experiments reported below, we use $m=100$, which is not \textit{very} extreme
but suffices for investigating the propensity models.
Each label $j$ is represented by a $d$-dimensional hyper-ball $S_j$,
whose radius and center are generated randomly.
All those hyper-balls lay in a feature space being itself a $d$-dimensional hyper-ball $S$,
big enough to contain all hyper-spheres $S_j$.
Instances are uniformly generated in $S$ 
and each instance $\sinstance$ is associated with labels 
whose hyper-balls contain $\sinstance$, i.e., $\relevantlabels{\sinstance} = \set{j \in [\numlabels]: \sinstance \in S_j}$.
The popularity (or priors) of the labels is directly determined by the radius of $S_j$.
We separately generate training, validation, and test sets from the above model.
We then apply a propensity model of choice to the training set to generate missing labels.
For some experiments, we also generate missing labels in validation and test sets.

For the original XMLC benchmark datasets, we assume that there are no missing labels. 
We then merge the original train and test sets, 
and take labels having at least $s$ positive instances.
We perform this step to select labels for which one can apply the noise models
without removing all positive labels. 
We then split the data again into training and test sets,
and apply, similarly as above, a propensity model of choice to the training set to generate missing labels. 
For some experiments, we extract a validation set from the test set.

In the original Yahoo R3 dataset, records are organized in a format of user-item ratings, 
and each record contains a user ID, an item ID, and the user's rating for the item (from 1 to 5). 
The training set contains over 300K ratings from 15.4K users to 1K items. 
This set is biased as users select items from a limited list of options recommended by some algorithm.
The bias-controlled test set is obtained by collecting ratings from a subset of 5.4K users 
to rate $r = 10$ randomly selected items. 
To create a multi-label dataset we treat each item as a label ($\numlabels=1K$).
We consider ratings greater or equal to 4 as positive feedback and others as irrelevant. 
We take users unique to the original training set (10K of users) to create a biased multi-label training set. 
For each user, we randomly split positive feedback into equal halves. 
We take the first half as features $\sinstance$ and the later half as labels $\slabels$. 
Next, we use users present in both the original training and test set to create a test set with a controlled bias. 
We again randomly select half of the positive feedbacks from the training set as features $\sinstance$ (to keep the same distribution of features) 
and all positive feedback from the test set as labels $\slabels$.
We can also extract a validation set from the test set,
usually consisting of a half of users from the test set. 

For a dataset created in such a way, 
we can calculate estimates of training-set propensities $\trainprop_j$ as:
\begin{equation}
    \label{eq:unbiased-propensities}
    \estimtrainprop_j = \phi_{\text{direct}} = \trainlabelprior_j \ctprop_j \left (\vallabelprior_j \right )^{-1}
    \,,
\end{equation}
where $\trainlabelprior_j$ and $\vallabelprior_j$ are, respectively, 
training- and validation-set estimates of the prior probability of label $j$, 
and $\ctprop_j$ is the controlled propensity used for the validation and test set.
To estimate the label priors, one can use relative frequencies of labels in training and validation sets.
For $\ctprop_j$ we use a ratio of $r$ labels used for labelling to all $m$ labels, i.e., $\ctprop_j = \frac{r}{\numlabels} = 0.01$
for the Yahoo R3 dataset. 

\subsection{Alternative propensity models}
\label{sec:alternative-propensity-models}

The \xmlcpropensitymodel{} propensity model~(\ref{eq:jain-prop-model}) is not
the only one to consider. In fact, many different forms have been introduced
in other domains~\citep{Saito_et_al_2020,Yang_et_al_2018,Joachims_et_al_2018}. 
We express the propensity models as functions of observed label priors
$\olabelprior_j \coloneqq \probability[\rolabel_j = 1]$, without direct relation to $n$, 
which by construction avoids convergence issues discussed in Section~\ref{subsec:shortcomings}.%
\footnote{%
The simplest estimate of the priors are relative frequencies of labels, i.e.,
$\hnoisylabelprior_j = \ocount_j/\numinstances$. As we deal with many very
sparse labels, we should rather use more robust estimates, for example,
$\hnoisylabelprior_j = (\ocount_j + \alpha)/(\numinstances + \alpha)$.%
}

A propensity model used frequently in recommendation systems is given by the following power-law formulation:
\begin{equation}
\propensity_j = \phipower(\olabelprior_j;\beta,\gamma) \coloneqq \left (\beta \olabelprior_j \right )^\gamma\,.
\end{equation}
With $\beta = \max_j \ocount_j/\numinstances$ we receive a model used, 
for example, in~\citep{Yang_et_al_2018, Saito_et_al_2020},
while for $\beta = \gamma= 1$ we get a very simple model which might be used,
if estimation of the parameters is infeasible due to lack of unbiased or bias-controlled data.
Another solution could be to use the generalized logistic function, also called Richard's curve~\citep{Richards_1959_curve},
which is very flexible and its shape resembles~(\ref{eq:jain-prop-model}):
\begin{equation}
\propensity_j = \phirichards(\olabelprior_j; c,d,e,f,g,h) \coloneqq c+\frac{d - c}{\left (e+ f\exp(-g \olabelprior_j) \right )^{1/h} }\,.
\end{equation}

The parameters of the models can be either set up according to a domain
knowledge or fit using an additional unbiased or bias-controlled dataset. In
the latter case, one can use standard non-linear optimization
methods~\citep{Bazaraa_nonlinear_programming}. We fit the models to inverse
propensities, minimizing squared errors $\|\estimprop_j^{-1} -
\phi(\olabelprior_j)^{-1}\|^2$ using the Levenberg-Marquardt
method~\citep{More_and_Watson_least_squares}. The errors are reported in the Appendix.

\input{figures/yahoo-r3-ps}

\autoref{fig:yahoo-r3-ps} illustrates the results of fitting the different propensity models for the
Yahoo R3 dataset. The actual training-set propensities have been obtained
using~(\ref{eq:unbiased-propensities}). The plot clearly shows that the
\xmlcpropensitymodel{} model with $a=0.55$ and $b=1.5$, suggested as default
values, is not a good fit to the actual propensities. The same model, but with
$a$ and $b$ fitted to the data gives a degenerated solution because many values are out of codomain of~(\ref{eq:jain-prop-model}) when $n$ is that small. 
On the other hand, $\phipower$ and $\phirichards$ seem to give a good fit, but still
actual propensities are widely spread, suggesting that a model solely
depending on label priors might not be the best choice.

We have also trained prediction models using the above propensities to see 
whether they help in improving (actual) precision$@k$ on the unbiased test set.
We use the one-vs-all approach in which probabilistic model $f_j(\sinstance)$, for label $j$, 
is obtained by minimizing the unbiased variant of logistic loss~\citep{Saito_et_al_2020, Qaraei_et_al_2021}:
\begin{equation}
\label{eq:un-log-loss}
\lossfn(\solabel_j, \propensity_j, f_j(\sinstance)) =
    -\frac{\solabel_j}{\propensity_j} \log(f_j(\sinstance)) - \left(1 -\frac{\solabel_j}{\propensity_j}\right) \log(1 -  f_j(\sinstance))\,.
\end{equation}

\input{tables/results-yahoo-r3}

The results are given in \autoref{tab:results-yahoo-r3}. 
As a baseline we also use a vanilla logistic loss which corresponds to $\phi_{1}(\olabelprior_j) = 1$.
We can observe that all propensities models, 
except the degenerated variant of \xmlcpropensitymodel{},
give slightly better results than the baseline, with $\phirichards$ being clearly the best among them. 
On the other hand, $\phi_{\text{direct}}$, 
which directly estimates propensity for each label using~(\ref{eq:unbiased-propensities}),
significantly improves the performance (particularly for $\precision[3]$ and $\precision[5]$).
Nevertheless, $\phi_{\text{direct}}$ can only work well if the unbiased or bias-controlled data are substantial. 
If this is not the case, one might need to use a parametric model, 
but the above results suggest 
that the dependence on label priors might not be sufficient.

\input{tables/results-artificial-data}
Finally, we illustrate a problem of propensity mismatch on synthetic and modified benchmark datasets.
We introduce noise to training data according to either the $\phijain$ or $\phipower$ model,
train prediction functions using both propensities models, and report actual precision$@k$, 
computed on the unbiased test set, along with propensity-based precision$@k$ for the same $\phijain$ or $\phipower$ model, 
computed on the biased test set (i.e., with the noise model applied).
The results in \autoref{tab:cross-propensities-p@k} show that relying on propensity-based metrics can be misleading.
As it should be expected, in the majority of cases, models compatible with the metric are obtaining the best performance.
However, selecting a model based on a chosen propensity-based metric can be wrong as the actual precision 
might be driven by a completely different propensity model.

\subsection{Propensity estimation via joint learning}
\label{sec:ps-joint-learning}

To minimize an unbiased loss function, such as the unbiased logistic
loss~(\ref{eq:un-log-loss}), one needs to know propensities in advance. 
However, estimating them might be difficult in practice. 
As demonstrated above, the use of inaccurate estimates can lead to results being far away from the optimal ones.

Therefore it would be useful if propensities could be estimated directly from a biased training set. 
Unfortunately, this is an ill-defined problem because the absence of a label can be explained by 
either a small conditional probability of the label or a low propensity or both. 
The additional assumption needed for the propensity to be identifiable were studied before, 
in the areas of learning from positive and unlabeled data~\citep{elkan_learning_2008}, 
and novelty detection~\citep{Blanchard_et_al_2010_SSND}.
The overview of the possible assumption is given by~\citet{Bekker_and_Davis_2020_PAU_Survey}, 
where the weakest of the assumptions requires 
that the true distribution of negative samples for a given label 
cannot contain the positive distribution~\citep{Blanchard_et_al_2010_SSND}. 
In these areas and under compatible assumptions, many methods for estimating the error ratio or labels priors, 
both directly related to propensity estimates, were proposed~\citep{Bekker_and_Davis_2020_PAU_Survey}.
Recently, \citep{Zhu_et_al_2020} and~\citep{Teisseyre_et_al_2020} have introduced methods for estimating the unbiased conditional label probabilities and propensities jointly on the biased training set.  
We refer to such methods as Propensity Estimation via Joint Learning (PEJL).

Let us briefly describe the method of \citet{Teisseyre_et_al_2020} 
(cf. Appendix for description of the method of \citet{Zhu_et_al_2020}).
It uses the fact 
that minimization of logistic loss leads to estimation of the posterior probability. 
Therefore, we can define the loss in the following way:
\begin{equation}
\label{eq:log-loss-ps-plug}
\lossfn(\solabel_j, \propensity_j, f_j(\sinstance)) \!=\!
    -\solabel_j \log(\propensity_j f_j(\sinstance)) \!-\! \left(1 \!-\!\solabel_j\right) \log(1\!-\!\propensity_j f_j(\sinstance))\,,
\end{equation}
where $\propensity_j f_j(\sinstance)$ can be seen as an estimate of the actual, 
ground-truth, conditional probability $\eta_j(\sinstance)$,
with $\propensity_j$ being the propensity and $f_j(\sinstance)$ the estimate of the observed conditional probability,
analogously to~(\ref{eqn:prob_adjustment}). 
This function can be optimized not only with respect to $f_j(\sinstance)$, but also to $\propensity_j$. The outline of the alternative method of \citet{Zhu_et_al_2020} can be found in the appendix.

We evaluate this approach on Yahoo R3 dataset. The estimated values of $\propensity_j$ are plotted on the \autoref{fig:yahoo-r3-ps} and the last row of \autoref{tab:results-yahoo-r3} presents the promising results of this approach. While the obtained estimates are overestimated, they capture the true trend. The predictive performance also looks promising, being only slightly worst than the best propensity model $\phirichards$. This is indeed encouraging as this method does not have access to the unbiased or bias-controlled data.
\autoref{fig:yahoo-r3-ps} also plots the obtained propensities for each label.

\subsection{Task losses for long-tails}
\label{sec:task-losses-for-long-tails}

It seems that \citet{Jain_et_al_2016} have introduced the propensity-scored losses 
rather to "promote" long-tail labels than to deal with missing labels. 
As such, the propensities can be seen as a kind of weighing approach that 
gives higher importance to less popular labels. 
Unfortunately, it is not clear why the weighing scheme used in~\citep{Jain_et_al_2016} should be preferred over 
other ones. 
Moreover, a weighing scheme does not have to be interpreted in terms of propensities.
Let us consider a weighted variant of $\precision[k]$:
\begin{equation}
\precision[k](\slabels, \spreds) = k^{-1} \sum_{\mathclap{j \in \topk[k](\spreds)}} w_j \solabel_j \,.
\end{equation}
This boils down to $\psprecision[k]$ when $w_j = \frac{1}{p_j}$ which also implies $w_j \ge 1$.
But one can use any weights that would represent the importance or gain of labels.
In such a case, the weighted $\precision[k]$ has a natural interpretation of being an unbiased estimate of the expected gain.
If tail labels are of our interest, then they should get higher weights, 
but actual values are rather domain-specific
without a direct relation to propensities.

To finalize our discussion, let us mention several other task losses that can be used 
as metrics that pay special attention to long-tail labels.
The macro $F_\beta$-measure defined as:
\begin{equation}
\fmacro[\beta] \left (\set{\slabels_i, \spreds_i}_1^\numinstances \right ) = 
\frac{1}{\numlabels} \sum_j \frac{(1+\beta^2) \sum_i \sblabel_{ij} \spred_{ij}} { \beta^2 \sum_i \sblabel_{ij} + \sum_i \spred_{ij}}\,,
\end{equation}
puts the same weight to each label and focuses on true positives. 
Therefore, positive predictions on long-tail labels are important to obtain high values on this metric.
One can also consider an $@k$ version of this metric, 
in which only top $k$ predictions are taken into account.

Another interesting metric, originally proposed for search engines~\citep{Radlinski_et_al_2008},
is abandonment$@k$ defined as:
\begin{equation}
\abandon[k](\slabels, \spreds) = \indicator[\forall j \in \topk[k](\spreds): \, \sblabel_j \ne 1 ]\,,
\end{equation}
which encounters no error if there is at least one correctly predicted label among the $k$ ones in the predicted set. 
This untypical formulation enforces diversity in the predicted set. 
Always predicting the two most popular but correlated, 
labels might be less beneficial than predicting less popular but also non-overlapping labels.

Finally, we mention the coverage metric, which directly reflects the diversity of correctly predicted labels.
It is defined as a fraction of labels with at least one correct positive prediction:
\begin{equation}
\!\!\!\coverage\!\left( \set{\slabels_i, \spreds_i}_1^\numinstances \right) = {\numlabels}^{-1} \left | \{ j \in [\numlabels] \!:\! \exists i \in [\numinstances] \; \text{s.t.}\; \sblabel_{ij} = \spred_{ij} = 1 \} \right | \,.
\end{equation}
This metric has already been suggested in the literature as an alternative~\citep{Babbar_Scholkopf_2019, Wei_and_Li_2020}.
It has also been used in the original paper of~\citet{Jain_et_al_2016},
but only to motivate the propensity-based metrics.

%% file: figures/yahoo-r3-ps.tex
\begin{figure}[t]
\centering
\includegraphics[width=0.75\linewidth]{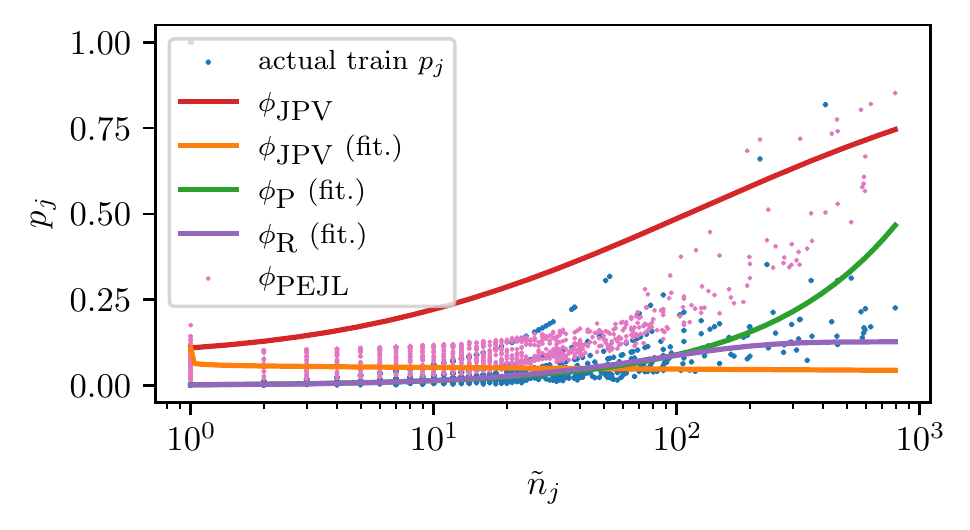}
\vspace{-0.4cm}
\caption{
Propensity models on the Yahoo R3 dataset. 
Annotation $\mathrm{(fit.)}$ 
denotes that the parameters have been fitted to the actual training-set propensities.
The $\phi_{PEJL}$ model is described in Section~\ref{sec:ps-joint-learning}.
}
\label{fig:yahoo-r3-ps}

\end{figure}

%% file: tables/results-yahoo-r3.tex
\begin{table}[t]
\caption{Actual precision@\{1,3,5\} (and their standard errors) on the Yahoo R3 dataset.
The best results are marked in bold.
The last row presents the results of the PEJL method from Section~\ref{sec:ps-joint-learning}. 
Each experiment was repeated 25 times.}
\label{tab:results-yahoo-r3}
\begin{center}
\small
\begin{tabular}{l|ccc}
\toprule
Method
& $\precision[1]$ (\%)
& $\precision[3]$ (\%)
& $\precision[5]$ (\%)
\\
\midrule
$\phi_{\mathbf{1}}$ & 60.83 $\pm$ 2.02 & 54.30 $\pm$ 0.89 & 51.20 $\pm$ 0.75 \\
$\phijpv$ & 66.03 $\pm$ 1.70 & 56.17 $\pm$ 0.91 & 52.02 $\pm$ 0.70 \\
$\phijpv$ (fit.) & 48.58 $\pm$ 2.13 & 43.26 $\pm$ 0.86 & 40.47 $\pm$ 0.68 \\
$\phipower$ (fit.) & 63.53 $\pm$ 1.90 & 54.53 $\pm$ 0.97 & 50.50 $\pm$ 0.73 \\
$\phirichards$ (fit.) & 71.23 $\pm$ 1.71 & 61.02 $\pm$ 0.77 & 54.03 $\pm$ 0.49 \\
$\phi_{\text{direct}}$ & \textbf{73.72 $\pm$ 2.26} & \textbf{66.14 $\pm$ 0.97} & \textbf{59.59 $\pm$ 0.74} \\
\midrule
$\phi_{\text{PEJL}}$ & 68.09 $\pm$ 1.53 & 58.15 $\pm$ 1.04 & 53.62 $\pm$ 0.72\\
\bottomrule
\end{tabular}
\end{center}
\end{table}

%% file: tables/results-artificial-data.tex
\definecolor{color-good}{HTML}{28a928}
\definecolor{color-wrong}{HTML}{df2021}

\begin{table}[b]
\caption{Mismatch of propensity models: actual P@$\{1,3,5\}$ (computed on unbiased test set)
and PSP@$1$ (computed on biased test set) of prediction models trained on data biased by 
$\phijpv$ or $\phipower$ models.
Green highlights PSP@k compatible with the used propensity model, while red highlights incompatible PSP@k.
The best value in each column for a given dataset is marked in bold.
}
\label{tab:cross-propensities-p@k}

\begin{center}
\resizebox{\linewidth}{!}{
\begin{tabular}{l|l|rrr|rr}
\toprule
Dataset
& Method
& \multicolumn{3}{c|}{P@k (\%)}
& \multicolumn{2}{c}{PSP@k(\%)}
\\
&
& @1
& @3
& @5
& ($\phijpv$) @1
& ($\phipower$) @1

\\
\midrule
\multirow{2}{*}{Artificial-$\phijpv$} & $\phijpv$ & \textbf{78.49} & \textbf{69.74} & \textbf{58.86} & {\color{color-good}\textbf{78.66}} & {\color{color-wrong}102.02} \\
 & $\phipower$ & 70.22 & 64.89 & 56.66 & {\color{color-good}70.34} & {\color{color-wrong}\textbf{133.60}} \\
\midrule
\multirow{2}{*}{Artificial-$\phipower$} & $\phijpv$ & 75.71 & 67.8 & 56.69 & {\color{color-wrong}\textbf{77.59}} & {\color{color-good}75.71} \\
 & $\phipower$ & \textbf{77.79} & \textbf{69.14} & \textbf{58.04} & {\color{color-wrong}72.02} & {\color{color-good}\textbf{77.20}} \\
\midrule
\multirow{2}{*}{EUR-Lex-$\phijpv$} & $\phijpv$ & 64.75 & 50.64 & 40.57 & {\color{color-good}64.94} & {\color{color-wrong}74.75} \\
 & $\phipower$ & \textbf{66.51} & \textbf{51.91} & \textbf{41.50} & {\color{color-good}\textbf{66.84}} & {\color{color-wrong}\textbf{80.63}} \\
\midrule
\multirow{2}{*}{EUR-Lex-$\phipower$} & $\phijpv$ & 54.23 & 41.72 & 32.99 & {\color{color-wrong}\textbf{44.33}} & {\color{color-good}52.19} \\
 & $\phipower$ & \textbf{55.07} & \textbf{42.07} & \textbf{33.04} & {\color{color-wrong}42.35} & {\color{color-good}\textbf{53.08}} \\
\midrule
\multirow{2}{*}{AmazonCat-$\phijpv$} & $\phijpv$ & \textbf{86.32} & 64.58 & 48.53 & {\color{color-good}\textbf{86.60}} & {\color{color-wrong}182.71} \\
 & $\phipower$ & 78.87 & \textbf{67.78} & \textbf{53.31} & {\color{color-good}79.38} & {\color{color-wrong}\textbf{389.35}} \\
\midrule
\multirow{2}{*}{AmazonCat-$\phipower$} & $\phijpv$ & 67.32 & 40.86 & 29.78 & {\color{color-wrong}\textbf{44.94}} & {\color{color-good}64.03} \\
 & $\phipower$ & \textbf{82.80} & \textbf{55.72} & \textbf{40.33} & {\color{color-wrong}44.31} & {\color{color-good}\textbf{82.22}} \\
\midrule
\multirow{2}{*}{Wiki10-$\phijpv$} & $\phijpv$ & \textbf{82.57} & \textbf{68.72} & \textbf{59.18} & {\color{color-good}\textbf{82.97}} & {\color{color-wrong}120.76} \\
 & $\phipower$ & 78.85 & 65.53 & 57.19 & {\color{color-good}80.01} & {\color{color-wrong}\textbf{228.02}} \\
\midrule
\multirow{2}{*}{Wiki10-$\phipower$} & $\phijpv$ & \textbf{80.44} & 54.8 & 47.06 & {\color{color-wrong}\textbf{87.34}} & {\color{color-good}\textbf{80.43}} \\
 & $\phipower$ & 79.18 & \textbf{59.89} & \textbf{49.05} & {\color{color-wrong}71.98} & {\color{color-good}78.61} \\
\bottomrule
\end{tabular}
}
\end{center}
\end{table}

%% file: 5-appendix.tex
\pagebreak

\section{Label frequency in XMLC datasets}

We show in \autoref{fig:labels-freqs} a log-log plot of the distribution of label frequencies
in popular benchmark datasets from the XMLC repository~\citep{Bhatia_et_al_2016}.
As also noted in other works~\citep{Bhatia_et_al_2015, Babbar_Scholkopf_2017},
the label frequencies are characterized by the long-tail. 

\input{figures/lables-freqs}

\section{Statistics of missing-label datasets}

We present in~\autoref{tab:new-dataset-stats} the statistics of datasets created to experiment with propensities models and missing labels. The description of these datasets is given in Section~\ref{sec:recipes-to-follow}. Because the process of generating the biased training sets contains randomness, for the mean number of labels per example, we report the average value from all generated variants of the datasets.

\begin{table}[h!]
\caption{Characteristics of datasets used in the experiments. 
We report the size of the \textit{biased} train set ($n^\text{tr.}$) and the size of the  test set ($n^\text{ts.}$), the total number of labels ($m$), and the mean number of labels per example in the \textit{biased} train set and the test set. Symbol $*$ denotes  the average value over all generated variants of the dataset, and $\dagger$ the value corrected by $p_j^c = r / m$, where $r$ is a number of labels sampled for labeling for each datapoint.}
\label{tab:new-dataset-stats}
\resizebox{\linewidth}{!}{
\begin{filecontents*}{data/datasets-new.txt}
{dataset}                   {n train}   {n test}    {m}     {avg l bs}      {avg l true}
Artificial-$\phijpv$     63000		30000		100		$\approx3.72^*$		4.27
Artificial-$\phipower$       63000		30000		100		$\approx2.44^*$		4.27
Eurlex-$\phijpv$         12188		5805		976		$\approx1.55^*$		4.46
Eurlex-$\phipower$           12188		5805		976		$\approx0.81^*$		4.46
AmazonCat-13K-$\phijpv$  627786		298946		1531	$\approx3.65^*$		4.43
AmazonCat-13K-$\phipower$    627786		298946		1331	$\approx0.69^*$	    4.43
Wiki10-31K-$\phijpv$     13079		6229		2951	$\approx6.22^*$		13.38
Wiki10-31K-$\phipower$       13079		6229		2951	$\approx1.46^*$	    13.38
Yahoo-R3                    11946		2436		1000	$\approx3.97^*$		$87.94^{\dagger}$
\end{filecontents*}
\pgfplotstabletypeset[
every head row/.style={
  before row={\toprule},
  after row={\midrule}},
  columns={dataset,n train,n test,m,avg l bs,avg l true},
every last row/.style={after row=\bottomrule},
columns/dataset/.style={string type, column type={l|}, column name={Dataset}},
columns/n train/.style={column name={$n^\text{tr.}$}},
columns/n test/.style={column name={$n^\text{ts.}$}},
columns/m/.style={column name={$m$}},
columns/avg l bs/.style={string type, column name={$\expectation[|\relevantlabels{\sinstance}|]^\text{tr.}$}},
columns/avg l true/.style={string type, column name={$\expectation[|\relevantlabels{\sinstance}|]^\text{ts.}$}}
]{data/datasets-new.txt}
}
\end{table}

\section{Estimation errors of propensity models}

\autoref{tab:est-error} presents the mean squared errors (MSE) of inverse propensity estimates ($\|\estimprop_j^{-1} - \phi(\olabelprior_j)^{-1}\|^2$) on the Yahoo R3 dataset
obtained by different propensity models described in Section~\ref{sec:recipes-to-follow}. Models marked as (fit.) have been fitted to the same data the error has been calculated on. 
Since we repeated the experiment with the PEJL model several times, we report the average error.

\begin{table}[h!]
    \caption{MSE on inverse propensity estimates of different propensity models for the Yahoo-R3 dataset. Symbol $*$ denotes the average value over several runs.}
    \label{tab:est-error}
    \centering
    \small
    \begin{tabular}{l|c}
        \toprule
        Model & MSE \\
        \midrule
        $\phi_{\mathbf{1}}$ & 1663.94\\
        $\phijpv$ & 1557.04 \\
        $\phijpv$ (fit.) & 1259.78 \\
        $\phipower$ (fit.) & 512.94 \\
        $\phirichards$ (fit.) & 516.85 \\
        \midrule
        $\phi_{\text{PEJL}}$ & $\approx1236^*$ \\
        \bottomrule
    \end{tabular}
\end{table}

\section{Details of model training}

We describe here the implementation and training procedures of classifiers used in the experiments in Section~\ref{sec:recipes-to-follow}.
For all the experiments, except the one with the PEJL model described in Section~\ref{sec:ps-joint-learning}, 
we train a linear model using the unbiased binary-cross entropy loss (\ref{eq:un-log-loss}) 
with propensities coming from different models $\phi$. 
The weights of the linear models are initialized from the uniform distribution $\mathcal{U}(-\sqrt{k}, \sqrt{k})$, 
where $k = 1/d$ with d being the number of features.

In the case of the PEJL approach, the same linear architecture is used but trained using~(\ref{eq:log-loss-ps-plug}). 
To assure that estimated values of $p_j$ are in $[0, 1]$, they are modeled as a sigmoid transformation ($\sigma(\cdot)$) of $p'_j$ parameters (one per label). Parameters $p'_j$ are initialized from the uniform distribution $\mathcal{U}(-e, e)$.

For each experiment, 10\% of the biased training set serves as a validation set for the selection of hyperparameters and early stopping. All methods are implemented using PyTorch~\citepappendix{pytorch}.
Optimization is performed with the Adam optimizer~\citepappendix{Kingma_et_al_2015}. 
Only two hyperparameters are tuned on the biased validation test, the learning rate (from set \{0.005, 0.01, 0.05, 0.1\}) and the weight decay (from set \{0, 1e-8, 1e-7, 1e-6\}). Experiments have been performed on a single machine with Intel Xeon Gold 5115 and NVIDIA V100 16GB.

\section{An alternative PEJL method}

Below we briefly describe the main idea behind the method of~\citet{Zhu_et_al_2020},
which can serve as an alternative to the approach of \citet{Teisseyre_et_al_2020}
describe in Section~\ref{sec:ps-joint-learning}.

Let $\masks \in \{0, 1\}^\numlabels$ be a mask random variable 
that expresses relation between $\rlabels$ and $\rolabels$, 
i.e., $\rolabels = \masks \odot \rlabels$.
We then have $\probability[\rolabel_j = 1 | \bx] = \probability[\mask_j = 1] \cdot \probability[\rblabel = 1 | \bx]$ 
and $\probability[\mask_j = 1] = p_j$ and $\probability[\rblabel = 1 | \bx] = \eta_j(\bx)$~\citep{Schultheis_Babbar_2021}. 
Because $\mask_j$ is a Bernoulli variable as well as $\rblabel_j$ and $\rolabel_j$, 
we end up with two models, the first one for $\rblabel_j$ with known $\propensity_j$, and the second for $\mask_j$ with known $\eta_j(\bx)$.
The first one can be obtained by minimizing~(\ref{eq:un-log-loss}). 
For the second one, a parametric model $\phi(\theta_j)$ for $p_j$ can be learned by minimizing the following logistic loss:
\begin{equation}
\label{eq:ps-un-log-loss}
\!\lossfn(\solabel_j, \eta_j(\bx), \phi(\theta_j)) \!=\! -\frac{\solabel_j}{\eta_j(\bx)} \log(\phi(\theta_j)) 
    - \left(1\!-\!\frac{\solabel_j}{\eta_j(\bx)}\right) \log(1 - \phi(\theta_j))\,.
\end{equation}
We can learn $f_j(\bx)$ and $\phi(\theta_j)$ jointly by replacing $\eta_j(\bx)$ by $f_j(\bx)$ in (\ref{eq:ps-un-log-loss}),
and $p_j$ by $\phi(\theta_j)$ in (\ref{eq:un-log-loss}).
Since both $f_j(\bx)$ and $\phi(\theta_j)$ can be updated on a single example $\bx$, to avoid estimation-training overlap problem, 
the training data is split into two parts, and the training is performed with $\phi(\theta_j)$ fixed on one part and with $f_j(\bx)$ fixed on the second one.

\section{Additional assumptions for complex metrics}

Here we show an example which demonstrates that for a non-decomposable metric, additional assumptions on the
process of labels going missing are required.

Consider a setting with two class labels. Let the label-wise propensities for both labels be $\propensity_1 = \propensity_2 = 0.5$.
The desired unbiased loss function of $\ell$ can be parametrized for each prediction $\spreds$ by four real numbers, $v_{\solabels} \coloneqq
\tilde{\ell}(\solabels, \spreds)$. We can consider two different scenarios. First, both labels always go missing at the
same time, second they go missing complementarily, corresponding to the probability distributions $\probability$ and
$\probability^\prime$. As the unbiasedness needs to hold for all potential distributions of true labels, it needs to hold in particular in the four cases in which the true label distribution is concentrated on a single point. By explicitly calculating expectations through reading off the probabilities from \autoref{tab:correlated-missing}, we can state the unbiasedness
requirement which is $\expectation[\tilde{\ell}(\rolabels, \rpreds)] = \expectation[\ell(\rlabels, \rpreds)] $:
\begin{equation}
\begin{aligned}
\ell((1,1), \spreds) &= 0.5 (v_{11} + v_{00}) = 0.5(v_{10} + v_{01})\,,\\
  \ell((1,0), \spreds) &= 0.5 v_{00} + 0.5 v_{10}\,,\\
  \ell((0,1), \spreds) &= 0.5 v_{00} + 0.5 v_{01}\,,\\
  \ell((0,0), \spreds) &= v_{00}\,.
\end{aligned}
\end{equation}
These are five linear equations with only four variables, so in general, there is no solution. If we additionally assume that
the labels go missing independently from each other, then the marginal propensities uniquely determine the full distribution.
In that case, unbiased estimates can be derived for general loss functions \citep{Schultheis_Babbar_2021}.

\begin{table}[h!] 
  \caption{Different distributions with the same propensities.}
  \label{tab:correlated-missing}
  \centering
  \small
  \begin{tabular}{cc|cc|cc|c}
    \toprule
    $\rblabel_1$ & $\rblabel_2$ & $\rolabel_1$ & $\rolabel_2$ & $\probability[\rolabel_1, \rolabel_2]$ & 
    $\probability^\prime[\rolabel_1, \rolabel_2]$ & $\tilde{\ell}(\solabels, \spreds)$ \\ \midrule
    1 & 1 & 1 & 1 & 0.5 & 0.0 & $v_{11}$ \\
    1 & 1 & 1 & 0 & 0.0 & 0.5 & $v_{10}$ \\
    1 & 1 & 0 & 1 & 0.0 & 0.5 & $v_{01}$ \\
    1 & 1 & 0 & 0 & 0.5 & 0.0 & $v_{00}$ \\ \midrule
    1 & 0 & 1 & 0 & 0.5 & 0.5 & $v_{10}$ \\
    1 & 0 & 0 & 0 & 0.5 & 0.5 & $v_{00}$ \\
    \midrule
    0 & 1 & 0 & 1 & 0.5 & 0.5 & $v_{01}$ \\
    0 & 1 & 0 & 0 & 0.5 & 0.5 & $v_{00}$ \\ \midrule
    0 & 0 & 0 & 0 & 1.0 & 1.0 & $v_{00}$ \\
    \bottomrule
  \end{tabular}
\end{table}

%% file: figures/lables-freqs.tex
\begin{figure}[h!]
\centering
\includegraphics[width=0.75\linewidth]{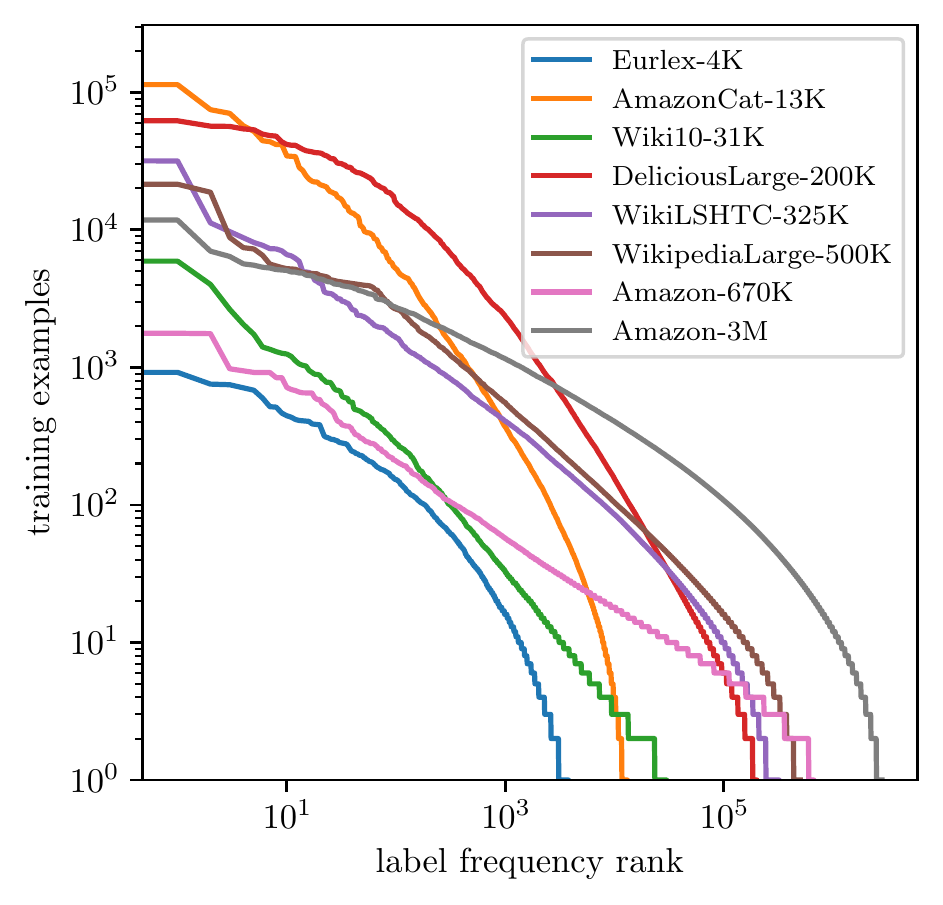}
\caption{
Label frequency in XMLC datasets. 
The X-axis shows the label rank when sorted by the frequency of positive instances 
and the Y-axis gives the number of the positive instances.}
\label{fig:labels-freqs}
\end{figure}